# A 3D MESH CONVOLUTION-BASED AUTOENCODER FOR GEOMETRY COMPRESSION

*Germain Bregeon*[1,2], *Marius Preda*[1], *Radu Ispas*[2], *Titus Zaharia*[1]

[1] Telecom SudParis, Institut Polytechnique de Paris, France
[2] Keyrus, Paris, France

## ABSTRACT

In this paper, we introduce a novel 3D mesh convolution-based autoencoder for geometry compression, able to deal with irregular mesh data without requiring neither preprocessing nor manifold/watertightness conditions. The proposed approach extracts meaningful latent representations by learning features directly from the mesh faces, while preserving connectivity through dedicated pooling and unpooling operations. The encoder compresses the input mesh into a compact base mesh space, which ensures that the latent space remains comparable. The decoder reconstructs the original connectivity and restores the compressed geometry to its full resolution. Extensive experiments on multi-class datasets demonstrate that our method outperforms state-of-the-art approaches in both 3D mesh geometry reconstruction and latent space classification tasks.
Code available at: github.com/germainGB/MeshConv3D

***Index Terms***— Mesh autoencoder, mesh geometry compression, mesh convolution, mesh reconstruction

## 1. INTRODUCTION

Autoencoders have gained significant success due to their ability to learn compact, meaningful representations of complex data through unsupervised learning. They excel in tasks such as data compression, denoising, and anomaly detection. However, adapting autoencoders to non-Euclidean data, such as 3D meshes, faces important challenges due to the irregular topological nature of such data. Several approaches have been explored to address this issue, including graph-based neural networks, spectral methods, and mesh-specific convolutions, each aiming to handle the complexity of mesh topology while learning efficient latent representations.

Most of the existing mesh autoencoders [1, 2, 3] are not very general. Indeed, they either impose a similar connectivity to all meshes by embedding the connectivity within the architecture, or the resulting latent space does not allow to directly compare different meshes. WrappingNet [4] is currently the only mesh autoencoder framework enabling general mesh unsupervised learning across objects with arbitrary connectivity. It exploits a low-resolution base graph, which facilitates learning a shared latent space representing the object's shape. This is achieved through a wrapping layer that wraps the base graph using latent features into the reconstructed mesh.

In this paper, we present a novel deep learning-based mesh autoencoder architecture for 3D geometry compression, building on the MeshConv3D representation introduced in [5]. Our model extracts features from each face and iteratively pools them while preserving connectivity through MeshConv3D operations, enabling efficient compression of mesh geometry. A dedicated reconstruction module then generates a base 3D mesh from low-resolution features, and a newly introduced unpooling layer reconstructs the connectivity of the base meshes back to the original connectivity. Unlike traditional compression methods that prioritize bitrate at the expense of downstream usability, our approach preserves a structured latent space that supports tasks such as classification, segmentation, and generation. It also satisfies the key conditions outlined in [4]: (1) it does not require input meshes to conform to a common template as in [1], and (2) it maintains a comparable latent space across meshes with varying topologies. Moreover, our solution can handle non-manifold and non-watertight meshes and supports multi-class representation learning, making it suitable for resource-constrained environments such as real-time 3D applications. While the current framework focuses on geometry compression, we are extending it to include connectivity compression, with the goal of developing comprehensive mesh-specific codecs. The main contributions of this paper are as follows:

- A refinement of the features introduced in MeshConv3D which allows to enrich the mesh representation,
- An unpooling process that makes it possible to reconstruct the initial mesh connectivity starting from a low resolution base mesh,
- An autoencoder architecture that relies on the pooling and unpooling operations to respectively define the encoder and decoder parts,
- A detailed experimental evaluation with results showing the superiority of our learned latent space notably for 3D mesh reconstruction tasks.

The rest of the paper is organized as follows. Section 2 provides an overview of the state of the art and analyzes the limitations of existing methods. Section 3 describes in details the proposed approach, with updated MeshConv3D operations, autoencoder architecture and reconstruction procedure. Section 4 summarizes the experimental results obtained on the SHREC11 [6] and Manifold40 [7] datasets. Finally, Section 5 concludes the paper and outlines directions for future work.

## 2. RELATED WORK

### 2.1. Deep Learning for 3D mesh feature extraction

In recent years, deep learning applied to 3D mesh analysis has seen significant advancements, particularly with respect to the issue of handling highly irregular topologies, characterized by varying neighborhood relationships.

Spectral methods such as DiffusionNet [2] provide a flexible solution: rather than enforcing a fixed mesh topology, they utilize the inherent spectral properties of the mesh to propagate information across vertices. To this purpose, DiffusionNet diffuse information through a multilayer perceptron neural network, which makes it possible to handle varying mesh structures without requiring explicit re-meshing.

Instead of operating directly on mesh vertices such as [8, 9], edge-based approaches define convolutions over the mesh edges. As representative of such approaches, let us cite the MeshCNN [10] model, where edge features are used for defining convolutional operations. The edge representation captures the connectivity between vertices and facilitates operations such as edge contraction for dimensionality reduction, thereby enhancing feature extraction while simplifying the data representation.

An alternative strategy relies on convolutional operations defined over mesh faces [7, 11, 12]. Within this framework, let us mention the SubdivNet [7] approach. Here, a Loop subdivision scheme is used to augment the mesh resolution. By densifying the mesh, pooling operations can be applied more effectively. On the counterpart, the subdivision process increases highly the number of faces and vertices, requiring significant computational resources for deep learning models.

### 2.2. Autoencoders for 3D Mesh Compression

Early methods treat the mesh as a graph. Subsequently, applying graph neural networks (GNNs) to learn representations becomes a natural solution. This is the case of the Variational Graph Autoencoders (VGAE) [13, 14], which use graph structures to encode and decode mesh information effectively. Unfortunately, the latent spaces of such methods can only be aligned for datasets with meshes that have the same connectivity and vertices ordered in the same way.

CoMA [1] is an autoencoder framework that employs hierarchical pooling based on quadratic error simplification, combined with spectral graph convolution layers. However, it operates under the constraint that all meshes in a dataset share the same size and connectivity, as both pooling and unpooling operations are defined on a fixed template graph.

The SAE approach [3] transforms the mesh data into the spectral domain and applies pattern-specific pooling to the spectral coefficients, enabling efficient feature extraction and compression. However, the spectral transformation requires here again a unique mesh connectivity.

CoSMA [15, 16] is an autoencoder designed to operate on subdivision meshes. Subdivision meshes, also known as semiregular meshes, have a hierarchical connectivity structure that enables efficient up/down-sampling using Loop subdivision and pooling [17]. Unlike CoMA, CoSMA does not require a shared template, making it applicable to mesh datasets with varying topology. The latent space of CoSMA consists of features associated to the faces of the base mesh. However, since the base mesh is not aligned across different meshes, the resulting latent spaces are not directly comparable.

WrappingNet [4] claims to be the first mesh autoencoder framework capable of processing meshes with varying topology while extracting a shared latent space comparable across all meshes. The original mesh is deformed into a canonical sphere that is used to transmit the connectivity of the mesh. The decoder uses then the latent representation to deform the sphere back into its reconstructed form.

The solution proposed in this paper is based on the MeshConv3D operators [5]. It enables us to create a network with properties comparable to those of WrappingNet [4]. Our architecture transforms the initial mesh space into a latent 3D base mesh space by reducing the number of faces and vertices from the input meshes. This allows the base meshes to remain distinguishable from one another. The key strength of our implementation lies in the mesh reconstruction module, which allows us to reconstruct meshes from the features extracted from each face of our meshes.

## 3. OVERVIEW OF THE METHOD

### 3.1. MeshConv3D

For the construction of our mesh autoencoder neural network, we extend the MeshConv3D approach that was previously introduced in [5], which provides deep learning operations applied on the mesh faces. The main functionalities provided are: feature extraction, pooling and unpooling of arbitrary 3D meshes. MeshConv3D operates directly on meshes with arbitrary topology, without requiring them to be watertight or manifold and without any need of re-meshing. In the following section, we provide an overview of the key MeshConv3D operations.

*3.1.1. Convolution operation*

In [5], a triangular mesh is defined as a set of vertices $\mathcal{V} = \{v_i | v_i \in \mathbb{R}^3\}$, a set of edges $\mathcal{E} = \{e_i | e_i \in \mathcal{V}^2\}$ connecting pairs of vertices, and a set of triangular faces $\{f_i | f_i \in \mathcal{V}^3\}$ expressed in terms of a connectivity matrix $\mathcal{F}$ of size $F \times 3$ storing the integer indices of the all mesh faces. The numbers of vertices, edges, and faces are denoted by $V$, $E$ and $F$ respectively.

MeshConv3D first defines a set of local patches for each face in the mesh, each containing $K$ faces, which determines the convolution region size and is user defined. A local patch includes the central face, its three direct neighbors, and additional neighboring faces selected by iteratively traversing adjacent faces until the patch reaches size $K$.

Then for a face $f$, we adapt the convolution operation from MeshConv3D to be defined as:

$$Conv(x_f) = W_0 x_f + W_1 \sum_{n=1}^{K} x_f^n + W_2 \sum_{n=1}^{K} |x_f - x_f^n| + W_3 \sum_{i=1, j>i}^{K} |x_f^i - x_f^j| \quad (1)$$

Here, $x_f$ represents the features vector of face $f$, while $x_f^n$, $n = 1, \ldots, K$ denote the features vectors of the other $K$ faces within the local patch of face $f$. The parameters $W_0$, $W_1$, $W_2$ and $W_3$ are learnable weight matrices of size (*fan*$_{out}$ × *fan*$_{in}$), with *fan*$_{in}$ and *fan*$_{out}$ respectively denoting the input and output feature dimensions of the convolution layer.

Compared to MeshConv3D's original convolution operation, we introduce here an additional component, controlled by the learnable weight matrix $W_3$. Its role is to measure differences between pairs of face features within the convolution region (excluding those of the central face) that are used to enrich the local patch representation.

As in MeshConv3D, we utilize sum operations, which are order-invariant and thus fundamental to MeshConv3D's convolution computation. This eliminates the need to define a specific order for the faces within local patches and represents a crucial property given the inherently non-Euclidean structure of meshes, where face ordering can be ambiguous. Furthermore, this operation enables a larger, user-defined receptive field.

*3.1.2. Pooling*

The face pooling operation (Fig. 1) first calculates a weight, denoted by $w_f$, for each face $f$ in the mesh. This weight is defined as the $L_2$ distance between the face's feature vector and those of its three neighboring triangles. It measures the local significance of the face within its neighborhood.

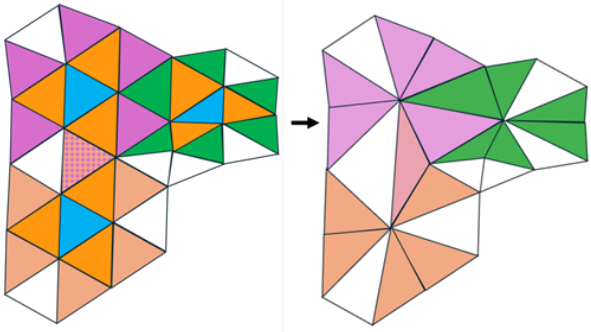

**Fig. 1.:** Pooling algorithm: the blue faces that correspond to the lowest $w_f$ weights are selected for pooling. The faces that share a comon edge with the blue faces (here in orange) are also removed. The white faces remain unaffected, whereas faces in other colors have their connectivity updated.

The entire pooling process is applied exclusively to the face features. The 3D vertex coordinates are never updated in this framework. The features of the faces whose connectivity is modified (represented in pink, beige, and green in Fig. 1) are defined as the average value of their neighboring faces within the corresponding pooling region.

The pooling algorithm iteratively selects groups of faces from each mesh for removal, prioritizing faces with the lowest weight, *i.e.*, faces that are located within the most uniform neighborhoods. Once selected, these faces are removed through a face pooling operation. The process continues until the meshes in the batch contain no more than $T$ faces, where $T$ is a user-defined parameter controlling the extent of the pooling operation.

The MeshConv3D pooling layer is highly efficient and well-suited for learning size reduction in meshes of arbitrary topology. We have adapted this pooling method to align with our new unpooling process, by storing at each pooling layer, a list of the faces that were removed (blue and orange faces in Fig. 1) along with the pooling region they originally belonged to (regions defined by the pink, beige, and green areas in Fig. 1). We also include a list of the faces in the pooled mesh whose connectivity was modified, as well as a corresponding list of the same size that contains the faces they were connected to but have been removed.

*3.1.3. Unpooling*

The unpooling operation consists of restoring, during the decoding process, the connectivity of the faces that were removed. Specifically, we reintegrate the removed faces based on their recorded positions, ensuring they are placed back correctly. The newly restored faces are initialized with the average feature values of their neighboring faces.

This process is repeated at each unpooling stage during decoding (Fig. 2) until the original mesh connectivity from the dataset is fully reconstructed.

**3.2. Autoencoder architecture**

*3.2.1. Input features to the network*

Our network operates on the mesh faces. Each face is initially represented by a feature vector formed by concatenating the geometric coordinates of its three vertices into a 9D vector, which is fed into the autoencoder.

*3.2.2. Overall architecture*

The autoencoder network consists of a series of convolutional blocks, each containing a convolutional layer, a batch normalization layer, and a ReLU activation function. In the encoder, these blocks are followed by pooling layers that progressively reduce the number of faces in the meshes.

Given the initial number of faces $F$ and vertices $V$, we strategically choose the target size $T$ for the base mesh. We reduce our meshes to a latent space of size $m$, requiring the geometry to be compressed to at most $\lfloor \frac{m}{3} \rfloor$ vertices. Achieving this reduction requires removing $T = (V - \lfloor m/3 \rfloor) \times 2$ faces from the original mesh, since each face pooling operation removes four faces but only two vertices. We propose the following scheme to define the pooling sizes ($T_1$, $T_2$ and $T_3$), aiming to reduce the mesh size uniformly while minimizing information loss during compression: $T_1 = F - T/3$ ; $T_2 = T_1 - T/3$ and $T_3 = T_2 - T/3$.

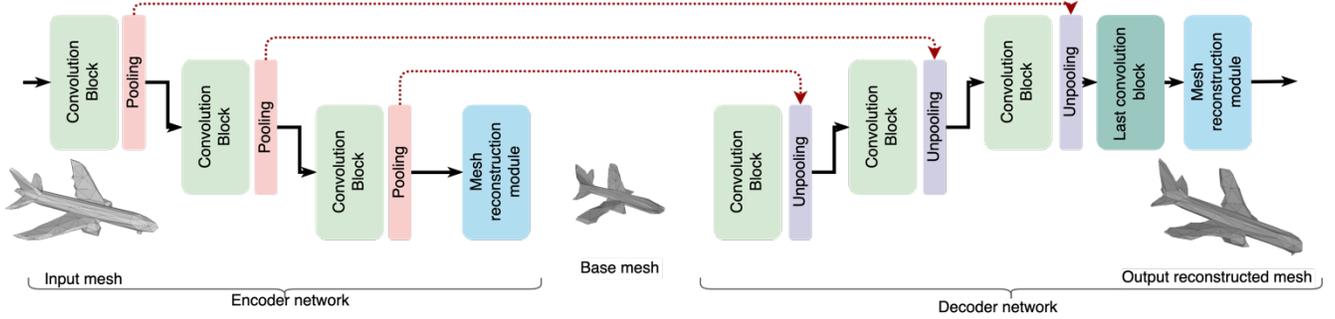

**Fig. 2.:** Overview of the architecture of our network. The original mesh to be compressed is fed into the network, passing through the convolutional blocks and pooling layers of the encoder network. The pooling layers transmit the connectivity they eliminate to the corresponding unpooling layers, enabling the reconstruction of the original mesh connectivity at the decoding stage. The two mesh reconstruction modules extract the learned features from the mesh faces and reconstruct the corresponding mesh geometry.

After the encoder's final pooling layer, the meshes are passed through the mesh reconstruction module, which reconstructs the base mesh forming our latent space.

The decoder mirrors the encoder's structure, applying unpooling layers to restore the mesh's connectivity, and convolution blocks to compute features on the faces of the meshes. Finally, after the last unpooling operation, a convolutional layer containing only the $W_0$ component of the convolution operation (1), meaning it acts solely on the central face without affecting its neighbors, ensures a final reduction of the number of features per face to 9. This guarantees compatibility with the input format required the mesh reconstruction module used to restore the mesh geometry and described in the following section.

*3.2.3. Mesh reconstruction module*

As a key component of our network, this module ensures that the learned latent feature space remains comparable across meshes with different connectivity, unlike most existing solutions to 3D mesh reconstruction problem. This allows meshes in the latent space to have varying connectivity while still being comparable within the 3D mesh space.

The reconstruction module processes the face features, which have been reduced to 9D vectors, and reconstructs the vertex coordinates of the meshes. The face connectivity matrix $\mathcal{F}$ is rebuilt at each level of detail during the encoding phase.

In the feature space representation, a given vertex is assigned a different 3D coordinate in each face to which it belongs to. We define its reconstructed 3D position as the average value of the corresponding 3D coordinates of all faces it belongs to. More precisely, for a given vertex $j$ its coordinates are computed as follows:

$$v_j = mean_i\{x_j(i) \mid \exists\, i \in \{0, \ldots, F\}, \exists\, k \in \{0,1,2\} \text{ such that } \mathcal{F}[i,k] = j\} \qquad (2)$$

where $x_j(i)$ represents the geometric coordinates (features) of vertex $j$ in face $i$.

In particular, this reconstruction process generates a complete 3D mesh in the latent space, from which (*cf.* §3.2.1) we extract the vertex positions of each face to serve as input features for the decoder. The same process is also applied at the end of the decoder pipeline to reconstruct the final mesh at the level of the decoder.

Let us now present the experimental evaluation of the proposed architecture.

## 4. EXPERIMENTAL RESULTS

Our study explores diverse mesh datasets featuring varying categories without relying on predefined templates. Each dataset consists of 2-manifold meshes. We assess the performance of newly built network on these datasets using the architecture illustrated in Fig. 2.

**4.1. Experimental setup**

*4.1.1. Datasets*

The experiments have been carried out on two widely used benchmark datasets: SHREC11 and Manifold40.

SHREC11 [6] consists of 30 classes, each containing 20 meshes with 500 faces per mesh. Following the experimental setup established in MeshCNN [10], we adopt the 16-split strategy, where 16 randomly selected models per class are used for training, while the remaining 4 serve as the test set.

Manifold40 [7] is composed exclusively of closed manifold meshes, each with 500 faces, derived from the ModelNet40 [18] dataset. It includes 40 object categories, with 9,843 meshes allocated for training and 2,468 for testing.

For both datasets, vertex positions are normalized to fit within a unit sphere. To enhance training, data augmentation techniques such as random scaling and random orientation are applied [7].

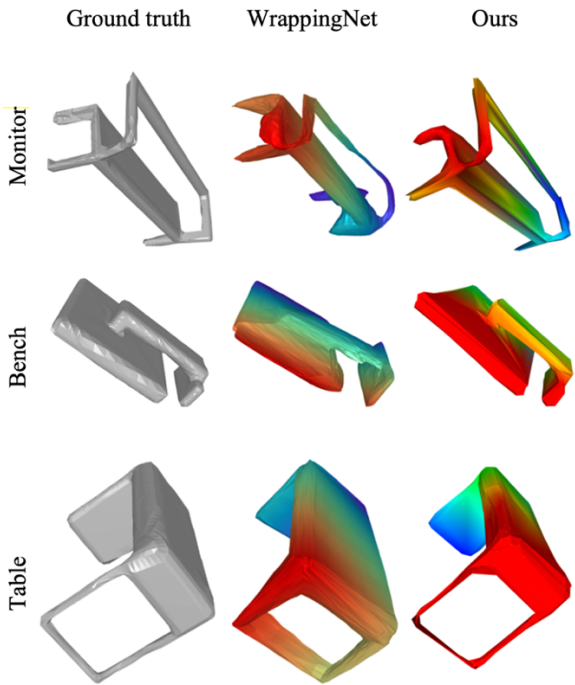

**Fig. 3.:** Qualitative comparison of reconstructed meshes

*4.1.2. Implementation*

We use Mean Squared Error (MSE) as the training loss, since the original connectivity is restored at decoding and vertex ordering remains consistent. Following the implementation of [4], we set the latent space size $m$ to 512 for this study and optimize the network using the Adam optimizer. We defined by grid search the output feature dimensions for our convolutional blocks in the encoder as [32,64,128] and in the decoder as [128,64,32].

As in [4], we compare our approach to point cloud autoencoders FoldingNet [19] and TearingNet [20], which extract fixed-length latent codes from variable-size point clouds, as well as WrappingNet [4] itself.

### 4.2. Shape classification in the latent space representation

To analyze our latent space, one approach is to solve a classification problem within it. We achieve this by generating latent space meshes using our encoder and then training a classification network on these meshes with the architecture introduced in MeshConv3D [5]. This evaluation demonstrates that the resulting meshes are well-suited for downstream deep learning.

Tab. 1 summarizes the classification results. Unlike our approach, which leverages a convolutional network, all other classification methods rely on an SVM, which may partly explain the performance differences. The accuracy, precision, and recall results for this classification task indicate that our method offers a significant gain in performances.

| | Manifold40 | | |
|---|---|---|---|
| **Method** | **Accuracy** | **Precision** | **Recall** |
| FoldingNet | 82.5% | .756 | .678 |
| TearingNet | 82.9% | .750 | .664 |
| WrappingNet | 83.3% | .778 | .730 |
| *Ours* | **89,8%** | **.857** | **.865** |

**Tab. 1.:** Classification results of the latent space of the different compared methods

### 4.3. Geometry compression task

We evaluate our method based on its geometric compression capabilities on databases. In this study, the compression of mesh connectivity is not taken into account.

| | SHREC11 | Manifold40 |
|---|---|---|
| **Method** | **CD / NE / CP** | **CD / NE / CP** |
| FoldingNet | .013 / .12 / .021 | 006 / .28 / .011 |
| TearingNet | .013 / .11 / .013 | .005 / .26 / .009 |
| WrappingNet | .023 / .09 / .004 | .005 / .22 / .004 |
| *Ours* | **.003 / .08 / .001** | **.004 / .16 / .002** |

**Tab. 2.:** Reconstruction metrics for the different methods on both datasets

Fig. 3 presents subjective results of mesh reconstruction using our autoencoder. We observe that the proposed approach has the ability to reconstruct fine mesh details, such as the legs of a table or a bench. In contrast, WrappingNet reconstructs the overall object's shape, but fails to preserve such fine details.

For objective evaluation, we follow the same methodology as WrappingNet. We compute the three metrics presented in [21]: Chamfer Distance (CD), Normal Error (NE), and Curvature Preservation (CP), and report their average values on the two testing sets retained (Tab. 2). The proposed method outperforms the other techniques across all metrics and datasets, with a particularly notable improvement on SHREC11, where the Chamfer Distance is reduced by a factor of 4 compared to FoldingNet and TearingNet, and by a factor of 7 compared to WrappingNet.

These results clearly demonstrate the superiority of our approach, which opens promising perspectives for geometry compression.

## 5. CONCLUSION AND PERSPECTIVES

In this paper, we have introduced a novel method for 3D mesh geometry compression. The proposed approach deals with meshes of varying sizes and connectivities, without imposing watertightness or manifold constraints. These features make our solution highly versatile for 3D mesh geometry compression. Furthermore, the experimental results demonstrate that it significantly outperforms state of the art techniques. For future work, we aim to integrate connectivity into the latent representation and further improve the mesh compression by jointly optimizing geometry and topology.